\documentclass[runningheads]{llncs}

\usepackage{eccv}

\usepackage{eccvabbrv}

\usepackage{graphicx}
\usepackage{booktabs}
\usepackage{amsmath,amssymb} 
\usepackage{algorithm}
\usepackage{algpseudocode}
\usepackage[accsupp]{axessibility}  
\usepackage{marvosym}

\usepackage{hyperref}

\usepackage{orcidlink}

\usepackage{pifont}

\newcommand{\xmark}{\ding{55}}
\usepackage{bm}
\usepackage{makecell}
\usepackage{xcolor}
\usepackage{makecell}
\usepackage{multirow}

\begin{document}

\title{Music-to-Dance Generation via \\Atomic Movements} 

\titlerunning{Music-to-Dance Generation via Atomic Movements}

\author{Xinhao Cai\inst{1}\orcidlink{0009-0009-5459-3458} \and
Yixuan Sun\inst{2}\and
Minghang Zheng\inst{1}\orcidlink{0000-0003-1612-975X} \and
Qingchao Chen\inst{3,4}\orcidlink{0000-0002-1216-5609} \and
Xin Jin\inst{4,5}\orcidlink{0000-0003-3873-1653} \and
Song-chun Zhu\inst{4,5,6}\orcidlink{0009-0009-9458-5583} \and
Yang Liu\inst{1,4}\textsuperscript{\Letter}\orcidlink{0000-0002-4259-3882}
}

\authorrunning{X.~Cai et al.}

\institute{Wangxuan Institute of Computer Technology, Peking University \and
School of Electronics Engineering and Computer Science, Peking University
\and
National Institute of Health Data Science, Peking University
\and
State Key Laboratory of General Artificial Intelligence, Peking University\\
\and
Beijing Institute for General Artificial Intelligence\\
\and
School of Intelligence Science and Technology, Peking University\\
\email{\{xinhao.cai, ilyssa\_syx\}@stu.pku.edu.cn} \\
\email{\{minghang,qingchao.chen,s.c.zhu,yangliu\}@pku.edu.cn}\\
\email{jinxinbesti@foxmail.com}\\
}

\maketitle

\begingroup
\renewcommand{\thefootnote}{}
\footnotetext{
\textsuperscript{\Letter} Corresponding author.}
\endgroup

\begin{abstract}
Music-driven dance generation aims to produce human motion that is both rhythmically synchronized and semantically consistent with music. While recent neural approaches have achieved impressive visual realism, they typically model motion as a continuous signal and neglect its compositional nature, making generated dances structurally incoherent and difficult to control.
In this work, we introduce a structure-aware framework that models choreography as a sequence of atomic movements—semantically interpretable motion events that serve as the building blocks of dance. To construct this atomic movement vocabulary, we first segment large-scale dance data and cluster them into atomic movement groups. We then employ a large language model to semantically relabel and refine the clusters, yielding a set of interpretable and reusable atomic movements.
Based on these atomic movement annotations, we design a two-stage generation framework that mirrors the human choreography process. In the atomic movement planning stage, the model predicts the type, duration, and timing of atomic movements conditioned on the input music, forming a symbolic dance allocation. In the completion stage, a transition-aware generator synthesizes smooth and stylistically coherent motion conditioned on the planned structure.
Extensive experiments demonstrate that our method produces dances with significantly improved structural coherence, rhythmic alignment, and perceptual naturalness compared to existing baselines, while providing enhanced interpretability and controllable editing through explicit structural representation. The code is available at https://github.com/oceanflowlab/AtomicDance.

\end{abstract}    
\begin{figure}[t]
\centering
\includegraphics[width=1\textwidth]{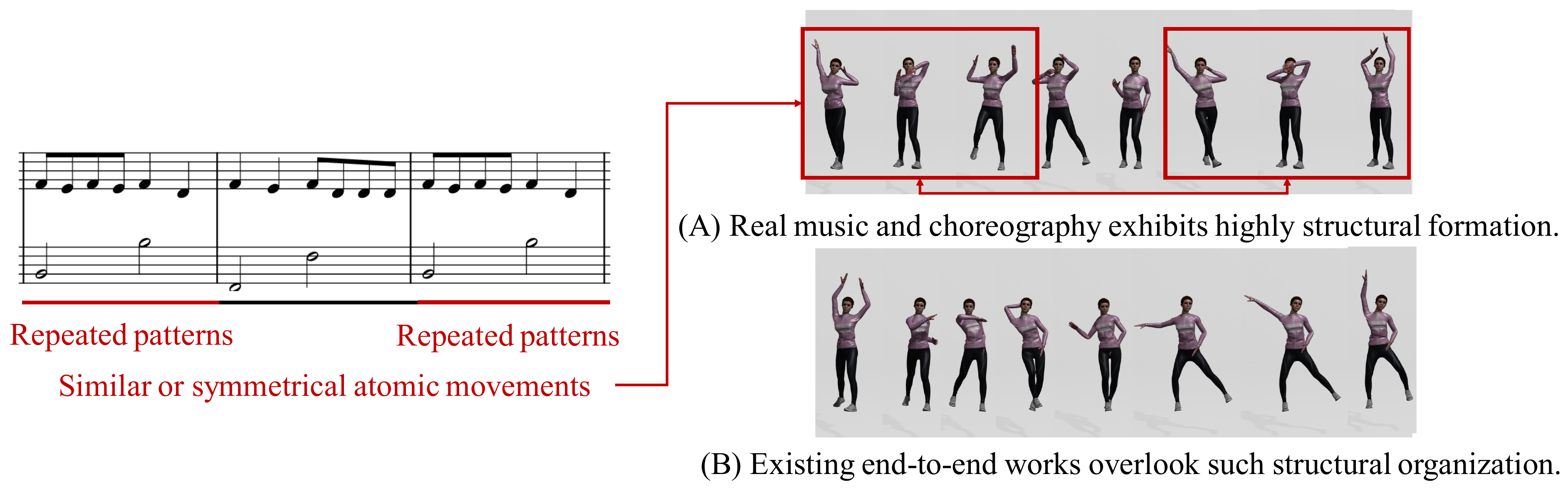}
\caption{\label{fig1} (A) There exists explicit repeated patterns in music and choreography. (B) Existing methods, usually in an end-to-end form, treat music-to-dance generation as a simple sequence-to-sequence task, neglecting the inner structure.
}
\end{figure}

\section{Introduction}
\label{sec:intro}

The task of generating dance conditioned on music has become an important research topic at the intersection of digital entertainment, embodied AI, and computational creativity. It aims to synthesize motions both physically natural and musically expressive, enabling applications in virtual performance, interactive art, and intelligent choreography.

Both choreography and music theory\cite{atmoic0,atomic1,atomic2,atmoic3,atomicstruct} emphasize that a performance is organized around discrete, meaningful, and reusable units. A choreographer first plans the sequence of such fundamental atomic movements and their temporal arrangement, while the dancer subsequently refines transitions and expressive variations.
However, most existing neural approaches\cite{edge} treat dance generation as a direct sequence-to-sequence translation. These models learn frame-level or short-term correspondences, such as beat alignment or local rhythm consistency, but overlook the \textbf{structural organization} that underlies real choreography. As a result, while they can produce visually smooth motion clips, the generated dances often lack segment-level coherence across sections. 

Thus, we revisit dance generation from a \textbf{structural and compositional} perspective. We hypothesize that complex dances can be represented as sequences of \textbf{atomic movements} that carry interpretable semantics and repeat or recombine to form longer compositions.
To realize this idea, we construct a high-quality set of \textbf{atomic movements}.
In this work, we define an atomic movement as a basic motion process that is repeatable, with variation, and can be freely combined. Specifically, an atomic movement must satisfy three criteria: 1) It must involve \textbf{a clear process} rather than a single frame. 
A static pose is comprehensive for a single moment but ultimately frozen in time. In contrast, an atomic movement unfolds through time, reveals shifts in emotion or energy, and allows more dynamic with the rhythm and phrasing of the music, thus creating a deeper artistic atmosphere. 2) It should have \textbf{repetition and variation}. The movements should reappear throughout the dance sequences and evolve in rhythm, dynamics, range, and direction. It may be sped up or slowed down, enlarged or minimized, reversed, fragmented, and reassembled. The coexistence of repetition and variation allows the movement vocabulary of the entire dance to remain both rich and diverse, yet unified. 3) It should be \textbf{semantically interpretable}, meaning that each atomic movement can be described naturally in language, which facilitates retrieval, and human-understandable labeling.

We then design a pipeline for constructing atomic motion groups.
First, we perform \textbf{segmentation} to divide continuous motion into procedural segments distinct from their surrounding context so that the movements involve clear processes.
Next, we apply \textbf{clustering} over motion features to identify frequently repeating motions across the dataset. The resulting clusters correspond to high-level and repetitive motion patterns, such as kicking or spinning, that capture the dominant movement tendencies of each motion, while preserving diverse variations within each group.
Finally, we perform \textbf{in-group re-clustering}: a tagging LLM annotates each motion segment with fine-grained natural-language descriptions, aligning segment semantics with human interpretation. A reasoning LLM then analyzes the annotations within each cluster to identify, summarize, and separate consistent intra-cluster patterns, re-clustering segments accordingly, yielding finer-grained, semantically interpretable motion categories.

Based on these learned atomic movements, we propose a two-stage structure-aware generation framework. In the \textbf{atomic movement planning} stage, given the input music, a planning network predicts which atomic movements should appear, along with their temporal locations and durations. This stage effectively produces a symbolic \textit{dance score}, describing the high-level choreography structure. In the \textbf{motion completion} stage, a diffusion-based generator synthesizes transitions for the planned atomic movements and re-create the chosen atomic movements for better diversity, ensuring smooth transitions and rhythmic synchronization. This two-stage design offers two advantages over conventional end-to-end approaches. First, it mirrors the real-world process of choreography: the first stage acts as the choreographer, planning the dance structure, while the second stage serves as the dancer, realizing the plan through expressive transitions. Second, because the atomic movements are discrete, meaningful, and temporally grounded, the generated dances become controllable—users can easily modify specific movements, adjust durations, or reassemble sections to produce new variations without retraining. These properties make our framework not only effective for generation but also practical for editing and interactive design.

Comprehensive experiments demonstrate that our method produces more structurally coherent, rhythmically aligned, and perceptually natural dances. Furthermore, the explicit atomic representation and two-stage pipeline enable unprecedented controllability and interpretability in dance generation.

Our main contributions are summarized as follows:
\begin{itemize}
    \item We propose an atomic movement discovery pipeline, which segments continuous motion into process-complete units, clusters them into repeating movement patterns while preserving inside-cluster variation, and clarifies finer-grained patterns via LLM-based re-clustering to produce semantically interpretable atomic movements. 
    \item We propose a two-stage structure-aware music-to-dance framework that separates atomic movement planning from transition-based motion completion, explicitly modeling choreography design and performance realization.
    \item Our approach achieves superior structural coherence, rhythmic consistency, and controllability compared to existing methods, offering a new perspective on interpretable and editable dance generation.
\end{itemize}

\section{Related Works}
\label{sec:rwork}

\subsection{Human Motion Generation}  
Recent advances in multimodal models\cite{gen1,gen2,mm1,mm2} have promoted growing interest in human motion generation. 

With the advent of deep generative models, condition-controlled motion generation became a major trend. TEMOS\cite{temos} and T2M-GPT\cite{t2mgpt} introduced VAE- or Transformer-based architectures, mapping free-form textual descriptions to 3D motions using large-scale datasets such as HumanML3D\cite{humanml3d}. Diffusion-based models, such as MDM\cite{mdm} and MotionDiffuse\cite{MD}, further improved sequential coherence and sample diversity. 
However, text-to-motion tasks provide concrete textual guidance and demand precise motion. In contrast, dance is an inherently artistic and compositional form that requires not only kinematic realism but also aesthetic creation, involving structural planning and artistic variation, where similar movements are creatively diversified. Therefore, directly applying text-conditioned motion frameworks to dance fails to capture these artistic and hierarchical characteristics. Our work instead focuses on uncovering and utilizing discrete atomic movements that serve as repeatable choreographic building blocks aligned with musical structure.

\subsection{Music-to-Dance Generation}  
Advances in generative models encourage the exploration of human motion generation across diverse scenarios, including human-object interaction\cite{interact}, while developments in pose estimation\cite{poseEst} and dance action recognition\cite{danceUnderstand} further promote music-to-dance generation, which aims to produce motion sequences that synchronize rhythmically and stylistically with music. With larger music-motion datasets, generative models became dominant. FACT\cite{aist} used a cross-modal Transformer to autoregressively predict joint trajectories from musical features. Bailando\cite{bailando} and TM2D\cite{tm2d} adopted VQ-VAE or GPT-based quantized representations to enhance music-conditioned controllability. Each discrete code in the learned vocabulary typically corresponds to a very short motion segment, resulting in decoded outputs that are essentially static poses and lack process information and segment-level structural information. EDGE\cite{edge} introduced diffusion models for dance generation with local editing and continuation capabilities. But these diffusion-based methods are usually trained within a short temporal window, which constrains their receptive field and prevents them from capturing global choreographic structures that unfold over longer timescales. Moreover, diffusion-based inpainting can edit motion at the signal level but lacks symbolic interpretability. In contrast, our pipeline handles the entire music at the first stage, therefore extracting more structural information of the music and is semantically interpretable at atomic movement level.

DanceBA\cite{danceba} and Beat-it\cite{beatit} optimize the alignment of rhythm. But they neglect the inner structure of the dance choreography. Later works consider long-sequence dance. LODGE\cite{lodge} proposed a coarse-to-fine diffusion pipeline for long dance sequences. EDMG\cite{edmg} further works on long dance generation efficiency.  However, they generate only keyframes without clear process and semantically interpretability, hindering the generation of a structural and compositional dance sequence conditioned on the music. 
Our approach complements these paradigms by introducing a structural layer of atomic actions, planned and synthesized hierarchically, bridging symbolic-level control and continuous motion fidelity.

\begin{figure*}[htb]
\centering
\includegraphics[width=0.72\textwidth]{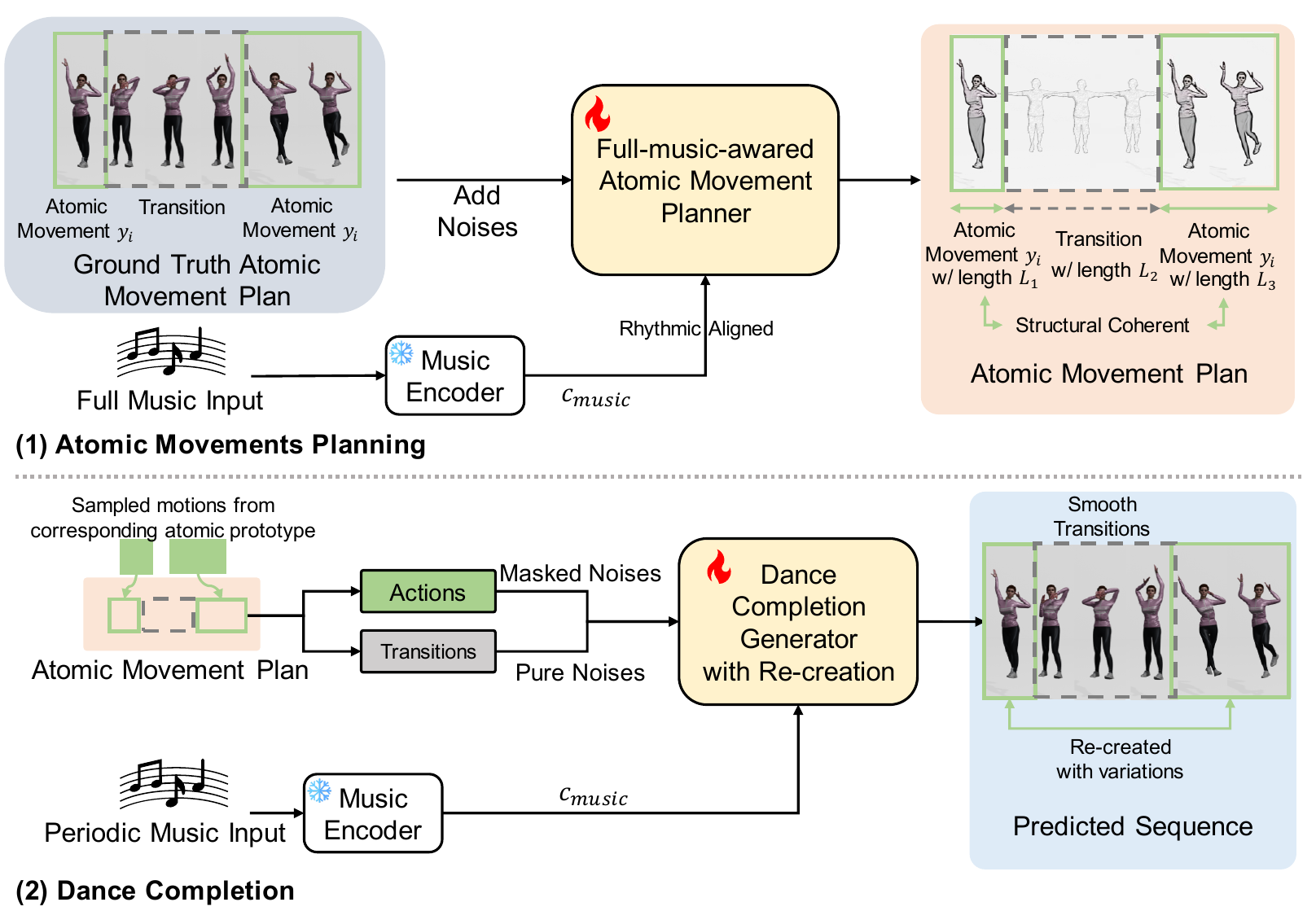}
\caption{\label{fig2}The framework of our method. We propose a two-stage pipeline. 1) Atomic movements planning: the Full-music-awared Atomic Movement Planner outputs the Atomic Movement Plan, allocating the atomic movement type, length and position. 2) Dance Completion: The Dance Completion Generator re-creates the atomic movements and generates the transitions between them. 
}
\end{figure*}

\section{Method}

\subsection{Problem Formulation and Overview}

Given a music condition $M^{1:L}$, where $L$ is the length of the music, our goal is to generate a dance sequence $X^{1:F}$ aligned with the music, where $F = L\times R_F$ and $R_F$ is a fixed frame rate. To enable the atomic-movement-based dance generation, we first collect and annotate the atomic movements included in the dance dataset, which we will introduce in \cref{sec::atomic anno}. 
Then, as we demonstrate in \cref{fig2} and \cref{sec::atomic plan}, our generation consists of two stages: (1) atomic movement planning and (2) dance completion. We first allocate the type, temporal position, and duration of each atomic movement conditioned on the full music. Subsequently, we optimize each movement and generate the transitions between them to make the dance continuous and fluent. 

\begin{figure*}[htb]
\centering
\includegraphics[width=1\textwidth]{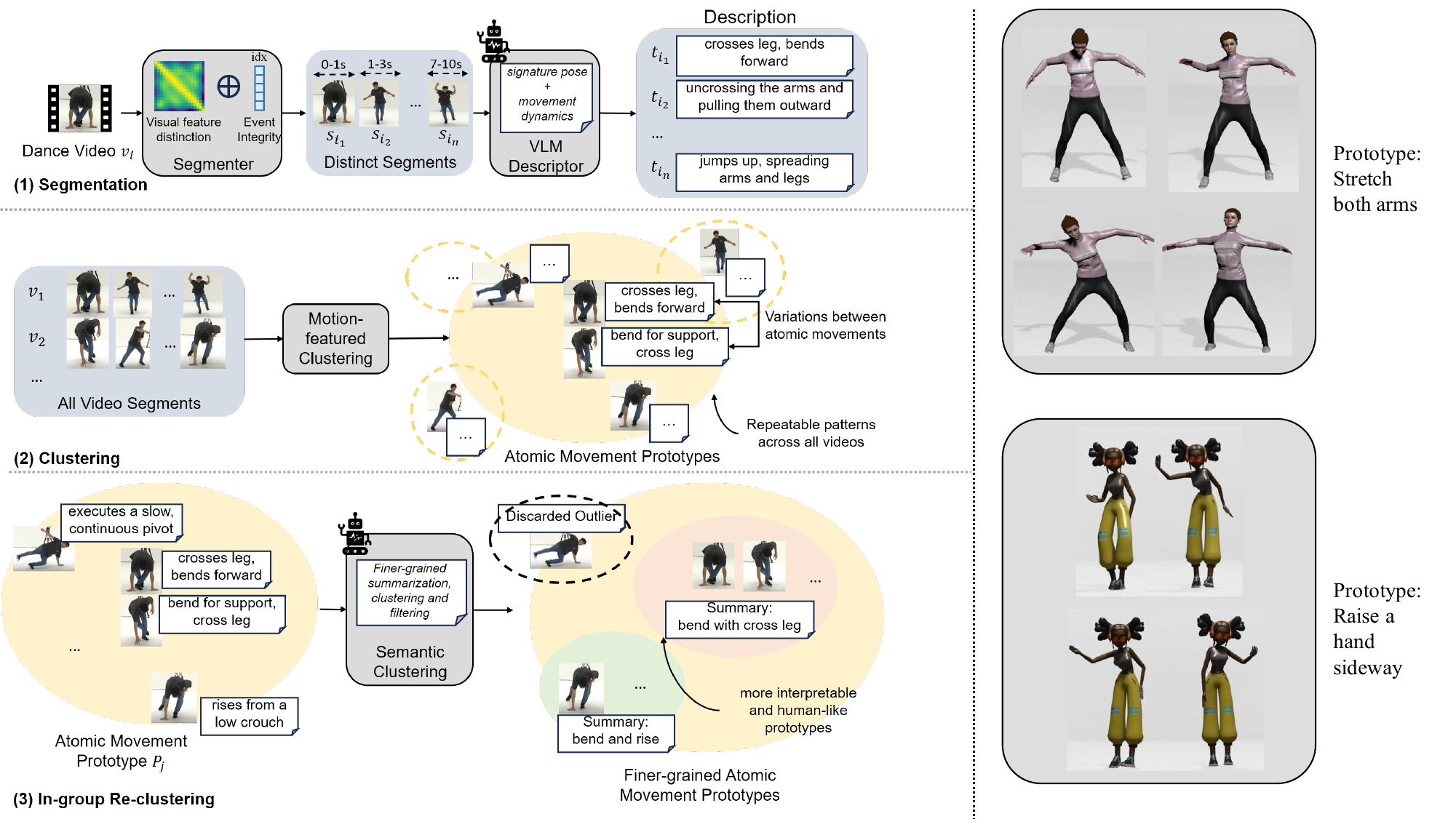}
\caption{\label{fig:cluster}The framework of our atomic movement discovery. 1) Segmentation exhibits distinct segments; 2) Clustering obtains atomic movement prototypes. 3) In-group Re-clustering re-splits each prototype by genres and leverages an LLM to further split them into finer-grained atomic movement prototypes. We show samples on the right.
}
\end{figure*}

\subsection{Atomic Movement Discovery}
\label{sec::atomic anno}

\paragraph{Pipeline Overview.} To find the atomic movements that 1) involve clear processes, 2) are repetitive and various, and 3) semantically interpretable, we design a three-stage pipeline to construct such atomic movement annotations(\cref{fig:cluster}) on AIST++\cite{aist} dataset. 
\textbf{(1) Segmentation}
: to obtain clear motion processes that are distinct from their temporal context, we employ a temporal-event-proposal method following \cite{nam2021zero} to segment continuous motion into complete, self-contained motion events. \textbf{(2) Clustering}: to identify basic motion units that repeats with variation, we encode each segment using TMR's \cite{petrovich23tmr} motion encoder and apply K-Means clustering across the entire dataset, organizing the motion segments into repeatable prototypes while preserving variation. \textbf{(3) In-group Re-clustering}: We use Gemini \cite{gemini} to provide fine-grained textual labels that align segment semantics with human interpretation, and a reasoning LLM abstracts and clarifies variation patterns within each cluster, producing finer-grained and semantically interpretable atomic movement categories.

\paragraph{Segmentation.}
We aim to extract complete motion processes that are clearly distinct from their temporal context. For example, a kick, which comprises the preparatory weight shift, leg extension, and recovery, forms a complete motion segment; any single frame of the extended leg or a very brief leg movement would fail to convey the full event. To obtain such complete motion processes, we follow the temporal event proposal method in \cite{nam2021zero}, performing segmentation by clustering by-frame similarity matrix and placing cut points at cluster boundaries. The idea is that frames are similar to frames within the same motion segment, and less similar to frames from other motion segments; therefore, frames within the same motion segment have the same similarity pattern, and tend to be clustered into a single motion segment. As detailed in \cref{alg:seg}, our segmentation algorithm partitions the dance sequence by clustering temporally-augmented visual similarity features. By identifying transitions in the cluster labels and iteratively merging segments shorter than $L_{min}$, the algorithm adaptively extracts motion units that preserve the structural integrity of the dance while responding to changes in visual patterns.

\begin{algorithm}[t]
\caption{Adaptive segmentation via visual similarity}
\begin{algorithmic}[1]
\label{alg:seg}
\Statex \textbf{Input:} Paired $T$-frame dance video $V = \{v_1, \ldots, v_T\}$ and $T$-frame 3D dance motion $M = \{m_1, \ldots, m_T\}$, $N$, minimum motion segment length $L_{\text{min}}$
\Statex \textbf{Output:} motion segment count $K$, cut points $B=\{b_k\}_{k=1}^{K}$ and motion units $S=\{s_i\}_{i=1}^{K+1}$.

\State Extract per-frame visual features $F = \{f_1,\ldots,f_T\}$ using an I3D encoder.
\State Compute a frame-wise similarity matrix $A\in\mathbb{R}^{T\times T}$ where
$A_{pq}=\mathrm{cos}(f_p,f_q)$.
\State $c_t \leftarrow A_{[t,:]} \in \mathbb {R}^{T}$ // Each row $t$ of $A$ encodes the similarity between frame $t$ and all other frames. 
\State $\tilde{c_t} \leftarrow [c_t; t/T]\in \mathbb{R}^{T+1}$ // Append normalized frame index to $c_t$ to encourage temporally local grouping.
\State Perform $N$-means clustering on $\{ \tilde{c_t}\}_{t=1}^T$ to obtain frame labels $\{l_t\}_{t=1}^T$.
\State Set cut points at label transitions:
$\tilde{B}=\{\,t \mid \ell_t \neq \ell_{t-1},\ 2\le t\le T\,\}$.
\State Iteratively merge segments shorter than $L_{\min}$ with the shorter of its temporal neighbors until all segments satisfy minimum length, yielding $B=\{b_k\}, K = |B|$.
\State Segment $M$ into motion units $S=\{s_i\}_{i=1}^{K+1}$, where
$s_i = M_{[b_{i-1},\, b_i)}$, with $b_0=1$ and $b_{K+1}=T+1$.
\end{algorithmic}
\end{algorithm}

Furthermore, we employ Gemini-2.5-Pro \cite{gemini} to describe each dance video segment in terms of a \textit{signature pose} and \textit{movement dynamics}. A signature pose is a relatively stable, sculptural posture that serves as a visual symbol of an atomic movement. Movement dynamics capture the transitions between static poses, encompassing the path, rhythm, intensity, and fluidity of motion, and thereby defining the diversity and specificity of each atomic movement. Specifically, we use PoseScript \cite{delmas2022posescript} to describe selected keyframes identified as \emph{motion beats} (local minima of segment-wise joint velocities). These PoseScript annotations are provided as auxiliary cues to the VLM descriptor, which then generates fine-grained natural language descriptions that capture both the signature pose and movement dynamics.

\begin{figure}[t]
  \centering
  \subfloat[]{\includegraphics[width=0.32\linewidth]{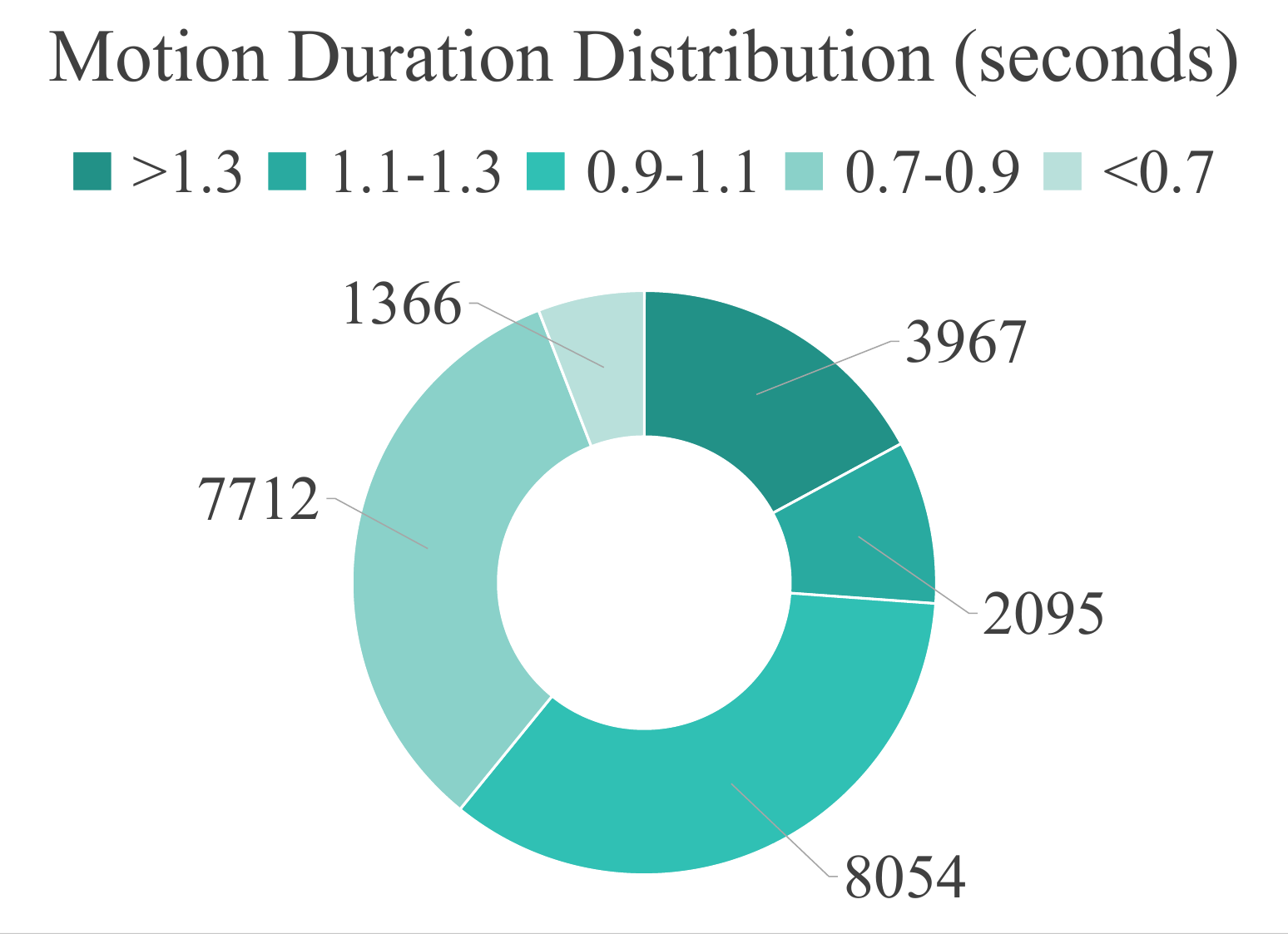}\label{fig:segmentlength}}
  \hfill
  \subfloat[]{\includegraphics[width=0.32\linewidth]{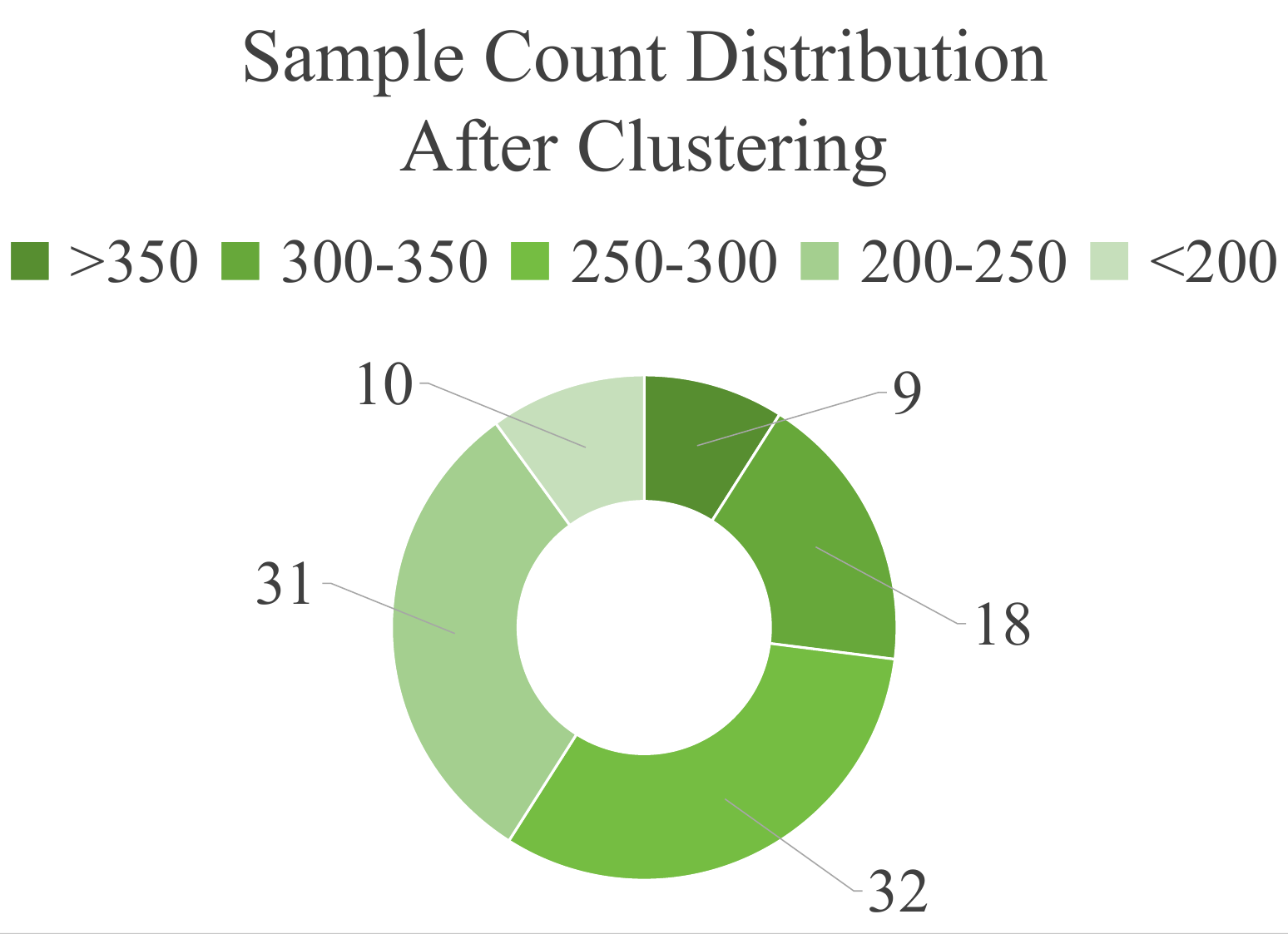}\label{fig:largecluster}}
  \hfill
  \subfloat[]{\includegraphics[width=0.32\linewidth]{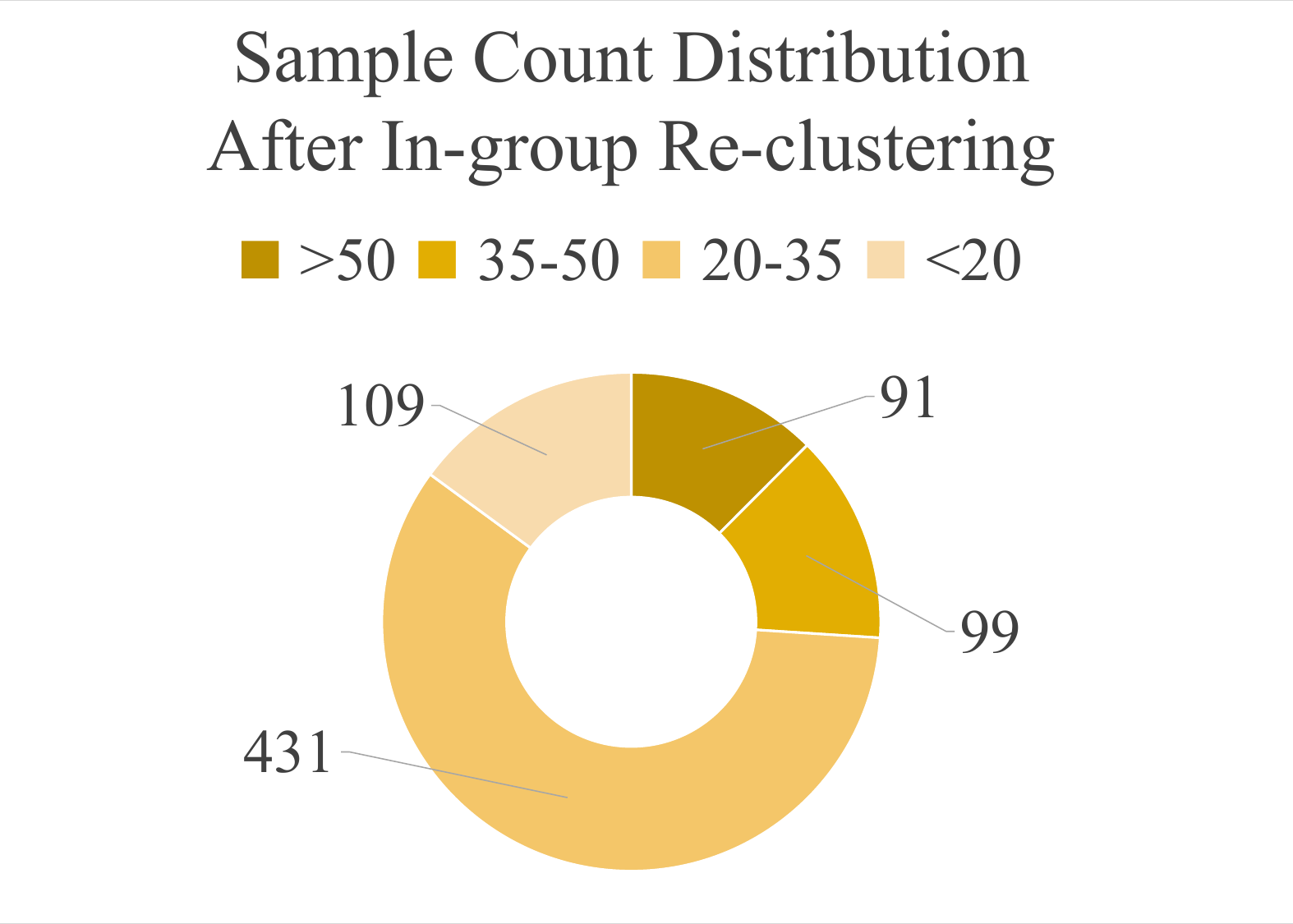}\label{fig:smallcluster}}
  \caption{Statistics of atomic motion lengths and clustering.
  } 
  \label{fig:abc}
\end{figure}

\paragraph{Clustering.}
To find the repetition and variation in atomic movements, we cluster motion segments across the entire dataset to extract atomic movement prototypes $\mathcal{P} = \{P_j\}$ that exhibit repeatability across pieces and preserve their variety.
Concretely, each segment is encoded by a pretrained TMR \cite{petrovich23tmr} motion encoder and projected into the joint motion-text embedding space; we then run K-Means to group segments into recurring atomic movement prototypes. TMR was trained with text-motion contrastive learning on HumanML3D \cite{humanml3d}, whose captions emphasize high-level action descriptions rather than fine-grained dance-specific details; as a result, the encoder naturally groups high-level similar segments while retaining intra-class variation. Because not every segment exhibits a clear pattern, we keep only segments near cluster centers and discard ambiguous edge points, yielding a set of high-confidence atomic movement prototypes. We yield 100 atomic movement prototypes from the complete AIST++ dataset; each prototype contains 268.57 segments on average (\cref{fig:largecluster}). This provides sufficient data density for the model to learn robust class representations. Also, these large clusters ensure that the atomic movements are repetitive, and the difference in a specific cluster guarantees the diversity.

\paragraph{In-group Re-clustering.}
To increase the interpretability of the atomic movements, we conduct semantical re-clustering for each prototype $P_j$. We further decompose each prototype to identify and label common patterns within a motion class. To better capture such intra-group patterns, we first pre-split each atomic movement prototype $P_j$ by dance genre, as motions from different genres tend to be semantically distinct.
To obtain finer-grained, more interpretable clusters, we use a summarizing LLM to iteratively: (i) identify a subset of mutually similar captions as a sub-prototype, and (ii) distill it into a concise semantic tag. We repeat this procedure until the number of ungrouped segments falls below a threshold. On average, each group yields 7.3 sub-prototypes with 31.8 samples each (\cref{fig:smallcluster}), preserving fine-grained diversity without severe data sparsity. This also makes the atomic movements more interpretable.

\subsection{Dance Generation}
\label{sec::atomic plan}
\textbf{Atomic Movement Planning.} Once a set of atomic movements has been defined, the central challenge is to determine how they should be temporally organized to form a complete dance that aligns with the given music and allocate these atomic movements to critical and appropriate positions. Traditional music-to-dance methods attempt to learn a direct mapping from music to dense motion frames, which captures local rhythm but ignores long-term compositional structure. In contrast, we explicitly model the planning process as predicting a symbolic sequence of atomic movements—each with a specific category, temporal position, and duration conditioned on full music instead of clips—thereby bridging musical phrasing with choreographic structure.

Given a music sequence, the planner aims to produce a symbolic \textit{dance score} that specifies when and how each atomic movement should occur. Our goal is to allocate every motion frame an atomic movement label or a token indicating that this frame should belong to a transition. We adopt a Discrete Denoising Diffusion Probabilistic Model (D3PM)\cite{D3PM}.

Assume that there are $K$ types of atomic movements, we use numbers $0$ to $K$ to denote these atomic movements and the transition. We have the atomic movement annotation for ground truth dance $Y_{gt} = [\hat{y_1},\hat{y_2},...,\hat{y_F}],y_i\in\{0,1,2,...,K\}$ denotes the atomic category of $i$-th frame or no atomic movement at $i$-th frame if $\hat{y_i} = 0$. Let $q(y_t | y_{t-1})$ denote the forward diffusion process that gradually perturbs the true atomic labels into random noise according to a transition matrix $Q_t \in \mathbb{R}^{(K+1)\times (K+1)}$:
\begin{equation}
q(y^t | y^{t-1}) = \text{Cat}(y^t; Q_t[y^{t-1}, :]),
\end{equation}
where $y^t$ is the atomic movement type at $t-$th frame and Cat stands for concatenating. At each step, a token is retained with probability $\alpha_t$ or replaced with a uniformly random category with probability $(1-\alpha_t)$. 
After $T$ steps, the sequence approaches a uniform distribution. 

We first use a pretrained music encoder to encode the music sequence $M$ into music features $c_{music} = \text{Enc}(M)$. The reverse process $p_\theta(y_{t-1} | y_t, \mathbf{c}_{music}, t)$ is parameterized by a Transformer-based diffusion model conditioned on the music features:
\begin{equation}
p_\theta(y_{t-1} | y_t, \mathbf{c}_{music}, t) = \text{Cat}(\pi_\theta(y_t, \mathbf{c}_{music}, t)).
\end{equation}
The model is trained by minimizing the variational objective:
\begin{equation}
\mathcal{L}_{\text{D3PM}} = - \mathbb{E}_q \big[ \log p_\theta(y_{t-1} | y_t, \mathbf{c}_{music}, t) \big].
\end{equation}

During inference, we start from a uniformly random sequence $y_T$ and iteratively sample from $p_\theta(y_{t-1} | y_t, \mathbf{c}_{music}, t)$ until $t=0$. 
The resulting sequence $\hat{\mathbf{Y}} = [\hat{y}_1, \dots, \hat{y}_F]$ represents the planned atomic movements, specifying both the category and temporal arrangement of each motion event. Also, we observed that any mislabeled frame within a continuous atomic movement segment could severely disrupt the planning results. Thus, we apply a lightweight post-processing module after the first-stage atomic movement planning. Specifically, we first perform a sliding-window majority vote to remove isolated frame-level errors and smooth local inconsistencies. We then apply a minimum-duration merging heuristic that detects abnormally short segments and reassigns them to the most semantically compatible neighboring segment. This two-step refinement effectively corrects over-segmentation errors while preserving genuine action boundaries, leading to more stable and coherent atomic-level plans.

This stage of atomic movement planning allows the model to allocate atomic movements to proper temporal positions and plan where to put the transitions between them, therefore improving the structural coherence and rhythmic alignment as the full music is used here. This explicit planning procedure also enhances the structural interpretability and segment-level controllability.

\textbf{Dance Completion.}
After obtaining the planned atomic movement sequence, the goal of the dance completion stage is to generate a continuous motion trajectory that connects these atomic segments smoothly while maintaining rhythmic and stylistic consistency with the given music.
Although the atomic movement sequence provides a structural outline of the choreography, it only specifies the categorical composition of movements. 
A realistic dance requires not only accurate realization of each atomic movement but also natural transitions between neighboring motions. Also, as choreographic theory\cite{atomicstruct} indicates, structured movements should exhibit variation when they recur. Accordingly, we apply appropriately scaled masked noise to the movement primitives to increase choreographic diversity and to improve their coherence with the preceding and succeeding transitional movements.
To this end, we introduce a diffusion-based completion model that refines and connects the planned motion sequence, ensuring continuity, musical alignment, and expressive diversity.

Given the predicted atomic label sequence $\hat{\mathbf{Y}} = [\hat{y}_1, \dots, \hat{y}_F]$, we first retrieve representative motion samples from the pre-clustered atomic movement database. 
For each atomic label $\hat{y}_i$, we select a motion segment whose duration most closely matches the planned temporal length, rescale it to fit the target duration, and place it into the corresponding frame range. For these chosen atomic movements, we will apply a masked noise, allowing them to exhibit better diversity and coherence with adjacent motions.
This yields a coarse motion sequence $\mathbf{M}_0$, in which atomic segments are filled while transitional frames remain uninitialized.
We employ a Denoising Diffusion Probabilistic Model(DDPM) \cite{ddpm} to refine $\mathbf{M}_0$ and generate smooth transitions. It consist of a forward perturbation process and a reverse denoising process. The former maps data $m_0$ to $m_t$ via $q(m_t | m) = \mathcal{N}(\sqrt{\alpha_t}m, (1 - \alpha_t)\mathbf{I})$, where $\alpha_t$ governs the noise schedule. During denoising, we optimize a network $f_\theta(m_t, t, \mathbf{x}_{music}, \mathbf{M}_0, \mathbf{w})$ to predict the clean signal $m_0$, with music condition $\mathbf{x}_{music}$, draft motion sequence $\mathbf{M}_0$ generated by stage1 and mask $\mathbf{w}$ controlling the noise ratio on the atomic movements chosen.

This masked strategy enables the model to preserve the semantics of known atomic movements while synthesizing diverse and musically aligned transitions. The training objective follows the standard denoising loss:
\begin{equation}
\mathcal{L}_{diff} = \mathbb{E}_{t,\mathbf{m}_0} \Big[ \big\| f_\theta(m_t, t, \mathbf{x}_{music}, \mathbf{M}_0, \mathbf{w}) - \mathbf{m}_0 \big\|_2^2 \Big].
\end{equation}
In addition, we apply a transition loss to ensure temporal smoothness across atomic boundaries:
$\mathcal{L}_{trans} = \sum_{b \in \text{boundaries}} \big\| \mathbf{M}_{b^-} - \mathbf{M}_{b^+} \big\|_1$,
which penalizes discontinuities between neighboring atomic movements, where $\mathbf{M}_{b^-}$ and $\mathbf{M}_{b^+}$ denote the pose frame before and after the atomic movement boundaries. The total training loss is thus:
$\mathcal{L} = \mathcal{L}_{diff} + \lambda_{trans}\mathcal{L}_{trans}$.
Consequently, $\hat{\mathbf{M}}$ exhibits coherent structure and smooth temporal dynamics.
To conclude, this dance completion stage generates transitions to make the dance more natural and smooth, while simultaneously allowing the atomic movement to be re-created to improve diversity and aesthetics.

\section{Experiments}
\label{sec:exp}

\subsection{Datasets and Implementation Details}
We conduct experiments on AIST++\cite{aist} dataset, which contains 1,408 music-paired 3D dance motion sequences, totaling 5.2 hours, covering 10 dance genres performed by 30 subjects with multi-view camera annotations. The atomic planner is trained as a music-conditioned discrete diffusion Transformer, while the completion module adopts a transition-aware continuous diffusion model based on the EDGE denoiser. Both modules use a latent dimension of $512$, $8$ Transformer layers, $8$ attention heads, feed-forward dimension $1024$, and dropout rate $0.1$. We train with AdamW using a learning rate of $2\times10^{-4}$, weight decay $0.01$, and gradient clipping of $1.0$.

\subsection{Metrics}
Following the existing music-to-dance generation works, we employ the following metrics to evaluate the quality of the dance sequences generated. 

\textbf{1) Frechet Inception Distance (FID).}
We compute the FID between the motion features of generated dances and those of the corresponding ground-truth sequences, qualifying the overall distribution distance in kinematic ($FID_k$) or geometry ($FID_g$) spaces. FID is lower the better. 
\textbf{2) Motion Diversity (Div).}
To assess the variety of generated dance motions, we follow \cite{bailando, lodge} and compute the mean pairwise Euclidean distance within the kinematic ($Div_k$) or geometry ($Div_g$) feature space. The closer Div is to the ground truth, the better it is. 
\textbf{3) Beat Alignment Score (BAS).}
To evaluate rhythmic synchronization between the generated dance and the input music, we adopt the Beat Alignment Score (BAS) as proposed in \cite{edge,lodge}. BAS is higher the better, To evaluate the structural consistency of the dance generated, we propose \textbf{4) $R$-precision.} We split the music into segments and calculate the music features of each clip. We examine whether the motion feature of the dances in every pair of the most-similar-music-clip is among the three highest most-similar-dance-clip. The precision is reported as the $R$. $R$ is higher the better. \textbf{5) MultiModality.} We report the variance between 5 samples generated by the same music as the MultiModality to test the diversity of the model. MultiModality is higher the better.

\begin{table*}[ht]
\centering
\caption{Quantitative results on AIST++\cite{aist}. $\uparrow$ indicates the performance is better if the value is higher. We use \textbf{bold} to represent the best result in the table.}
\scalebox{0.75}{
\begin{tabular}{c|cccccc}
    \toprule
    Method & $FID_k$$\downarrow$ & $FID_g$$\downarrow$ & $Div_k$ $\xrightarrow{}$ & $Div_g$ $\xrightarrow{}$ & BAS$\uparrow$ & $R$ $\uparrow$ \\
    \hline
    Ground Truth & 17.1 & 10.6 & 8.19 & 7.45 & 0.2374 & 42.1 \\ \hline
    DanceNet\cite{dancenet} & 69.18 & 25.49 & 2.8 & 2.85 & 0.143 & 14.1 \\ 
    DanceRevolution\cite{dancerevo} & 73.42 & 25.92 & 3.52 & 4.87 & 0.195 & 13.7 \\ 
    Bailando\cite{bailando} & 28.16 & 9.62 & 7.83 & 6.34 & 0.2332 & 17.9 \\ 
    EDGE\cite{edge} & 42.16 & 22.12 & 3.96 & 4.61 & 0.2334 & 16.3 \\ 
    Lodge\cite{lodge} & 37.09 & 18.79 & 5.58 & 4.85 & 0.2423 & 18.2 \\ 
    Ours & \textbf{25.26} & \textbf{9.03} & \textbf{8.01} & \textbf{6.69} & \textbf{0.2470} & \textbf{26.6} \\ 
    \bottomrule
\end{tabular}
}
\label{tab:exp1}
\end{table*}

\subsection{Quantitative Comparisons}
\Cref{tab:exp1} presents a comparison of our method against representative diffuison-based (EDGE\cite{edge}, Lodge\cite{lodge}) and VQVAE-based (Bailando\cite{bailando}) baselines. 

\textbf{1) In terms of motion fidelity}, our method achieves the lowest FID\textsubscript{k} and FID\textsubscript{g}, outperforming all diffusion- and VQ-VAE-based baselines. 
This indicates that the generated dances are not only physically realistic in terms of kinematic smoothness but also compositionally coherent in spatial structure. Such results prove that our method better captures the atomic movements, which are the common patterns shown in the real dance, and thus produce results that are closer to ground truth.
\textbf{2) For motion diversity}, our approach achieves high scores in both Div\textsubscript{k} and Div\textsubscript{g}, approaching the diversity level of real human dances.
This demonstrates that the proposed two-stage framework does not sacrifice diversity for structural regularity—on the contrary, by introducing partial noise during motion completion, our model enriches local variations within each atomic movement and generates a wider range of expressive dynamics.
\textbf{3) In terms of rhythmic alignment}, our model attains the highest BAS, indicating superior synchronization with musical beats.
This benefit arises from the explicit temporal planning in the first stage, which allocates atomic movements according to musical structure, and from the rhythm-conditioned diffusion refinement that further enhances temporal coherence.
\textbf{4) For structure consistency,}
our method achieves the highest $R$ score, substantially outperforming previous methods that lack explicit compositional modeling. This result confirms that our framework successfully captures the relationship between musical and choreographic structure—when two music segments share similar phrasing or rhythm, the generated dances exhibit correspondingly similar motion patterns.

Overall, these results quantitatively validate that our structure-aware two-stage framework produces more coherent, musically aligned, and controllable dances than existing end-to-end generative baselines.

\subsection{Ablation Studies and Discussions}

\textbf{Ablation of Atomic Movement Discovery.}
We first analyze the design of atomic movement discovery, which aims to construct meaningful and reusable atomic movement categories from motion data.
As shown in \cref{tab:cluster}, we compare different clustering configurations and investigate the effect of the proposed LLM-assisted re-clustering strategy.
When varying the base number of clusters from 75 to 125 without re-clustering, an excessively small number of clusters tends to over-merge distinct motion patterns, which weakens geometric fidelity and choreography quality.
In contrast, using too many clusters may over-fragment similar motions and reduce motion coherence.
These results indicate that atomic movement discovery requires a proper granularity to balance semantic consistency and motion diversity.

More importantly, introducing re-clustering improves the overall performance.
Compared with the naïve clustering strategy, the LLM-assisted semantic refinement achieves better FID scores, stronger beat alignment, and higher retrieval accuracy.
This demonstrates that language priors help merge semantically similar yet dynamically diverse movements into more interpretable atomic movement categories.
Therefore, the proposed atomic movement discovery process provides a more reliable foundation for subsequent planning and completion stages.

\begin{table}[t]
    \centering
    \caption{Ablation studies on atomic movement discovery.}
    \scalebox{0.75}{
    \begin{tabular}{cc|cccccc}
    \toprule
        \makecell[c]{Cluster \\Base Num} & ReCluster? &  $FID_k$$\downarrow$ &  $FID_g$$\downarrow$ & $Div_k$ $\xrightarrow{}$ & $Div_g$ $\xrightarrow{}$ & BAS$\uparrow$ & $R$ $\uparrow$ \\
    \hline
        75 & \xmark & 33.24 & 13.31 & 7.81 & 6.12 & 0.2413 & 20.2 \\ 
        100 & \xmark & 32.68 & 12.09 & 7.78 & 6.05 & 0.2420 & 21.3 \\  
        125 & \xmark & 34.57 & 14.28 & 7.75 & 6.01 & 0.2415 & 21.1 \\ 
        100 & \makecell[c]{w/o LLM} & 30.11 & 11.27 & 8.25 & 6.05 & 0.2431 & 23.3 \\ 
        100 & \makecell[c]{w/ LLM} & 25.26 & 9.03 & 8.01 & 6.69 & 0.2470 & 26.6 \\  
        \bottomrule
    \end{tabular}}
    \label{tab:cluster}
\end{table}

\begin{table*}[t]
\centering
\caption{Ablation studies on atomic movement planning.}
\scalebox{0.75}{
\begin{tabular}{c|cccccc}
    \toprule
    Method & $FID_k$$\downarrow$ & $FID_g$$\downarrow$ & $Div_k$ $\xrightarrow{}$ & $Div_g$ $\xrightarrow{}$ & BAS$\uparrow$ & $R$ $\uparrow$ \\
    \hline
    w/o Post-process & 25.26 & 9.03 & 8.01 & 6.69 & 0.2470 & 26.6 \\
    w/ Post-process & 24.02 & 8.78 & 8.03 & 6.68 & 0.2472 & 27.5 \\
    Using GT & 21.59 & 8.82 & 8.13 & 6.94 & 0.2455 & 29.8 \\
    \bottomrule
\end{tabular}
}
\label{tab:postpro}
\end{table*}

\textbf{Ablation of Atomic Movement Planning.}
We then evaluate the atomic movement planning stage, which predicts the temporal allocation of atomic movements before motion generation.
\Cref{tab:postpro} compares three settings: using planning results without post-processing, applying the proposed post-processing, and directly using ground-truth atomic labels.
(1) Without post-processing, frame-level mispredictions in the planner may fragment continuous atomic movements, which further introduces inconsistencies into the subsequent dance completion stage.
(2) After applying majority-vote smoothing and minimum-duration merging, both fidelity and alignment are improved, showing that corrected atomic boundaries lead to more stable and musically aligned motion generation.
(3) Using ground-truth atomic movement labels further improves the performance, demonstrating that more accurate atomic movement planning can directly benefit the final dance quality.
This also suggests that the proposed framework is compatible with human editing: if users manually adjust the planned atomic movement allocation before the completion stage, the system can potentially generate dances with better structure and quality.

\textbf{Ablation of Dance Completion.}
Finally, we study the dance completion stage, where the planned atomic movements are transformed into complete dance sequences.
We first examine whether allowing the re-creation of retrieved atomic movements contributes to overall motion quality.
In the \textit{Fixed Atomic Movements} setting, the selected atomic instances are directly concatenated and temporally rescaled without further modification.
In contrast, the \textit{Flexible Atomic Movements} setting allows the diffusion-based generator to adjust and refine each atomic movement during denoising, enabling smoother transitions and better adaptation to the accompanying music.
As reported in \cref{tab:e2e}, the fixed version exhibits lower motion diversity and weaker beat alignment, indicating that directly copying retrieved motion instances limits expressiveness and synchronization.
By enabling flexible completion, the model achieves a better balance between realism, rhythmic coherence, and diversity.
This also better matches real-world dance performance, where dancers naturally introduce variations when repeating similar atomic movements.
The MultiModality result further shows that flexible atomic movements lead to more diverse outputs under the same music, proving that our method does not simply copy movements from the dataset but adapts them into varied and coherent dance sequences.

\begin{table}[t]
    \centering
    \caption{Ablation studies on flexible atomic movement completion.}
    \scalebox{0.75}{
    \begin{tabular}{c|ccccccc}
    \toprule
        Method &  $FID_k$$\downarrow$ &  $FID_g$$\downarrow$ & $Div_k$ $\xrightarrow{}$ & $Div_g$ $\xrightarrow{}$ & BAS$\uparrow$ & $R$ $\uparrow$ & $MultiModality\uparrow$\\
    \hline
        Ground Truth & 17.1 & 10.6 & 8.19 & 7.45 & 0.2374 & 42.1 & -\\ \hline
        \makecell[c]{Fixed Atomic Movements} & 24.13 & 8.44 & 6.99 & 4.61 & 0.2356 & 27.2 & 1.43\\ 
        \makecell[c]{Flexible Atomic Movements} & 25.26 & 9.03 & 8.01 & 6.69 & 0.2470 & 26.6 & 2.25\\
        \bottomrule
    \end{tabular}}
    \label{tab:e2e}
\end{table}

\begin{table}[t]
    \centering
    \caption{Ablation studies on atomic movement retrieval for dance completion.}
    \scalebox{0.75}{
    \begin{tabular}{c|ccccccc}
    \toprule
        Method &  $FID_k$$\downarrow$ &  $FID_g$$\downarrow$ & $Div_k$ $\xrightarrow{}$ & $Div_g$ $\xrightarrow{}$ & BAS$\uparrow$ & $R$ $\uparrow$ & $MultiModality\uparrow$\\
    \hline
        Ground Truth & 17.1 & 10.6 & 8.19 & 7.45 & 0.2374 & 42.1 & -\\ \hline
        Single Candidate & 25.30 & 9.01 & 6.42 & 4.11 & 0.2458 & 28.9 & 2.18\\ 
        Random Choice & 27.94 & 9.98 & 8.07 & 6.73 & 0.2466 & 24.5 & 2.62 \\ 
        Duration Nearest & 25.26 & 9.03 & 8.01 & 6.69 & 0.2470 & 26.6 & 2.25 \\ 
        \bottomrule
    \end{tabular}}
    \label{tab:random}
\end{table}

We further investigate how different instance selection strategies affect dance completion when retrieving atomic movements from the discovered categories.
We compare three variants:
(1) \textit{Single Candidate}, where each atomic category is always represented by a fixed motion instance;
(2) \textit{Random Choice}, where an instance is randomly sampled from the corresponding category;
and (3) \textit{Duration Nearest}, where we select the instance whose length is closest to the predicted duration.
As shown in \cref{tab:random}, \textit{Single Candidate} produces the lowest motion diversity because using one fixed instance suppresses the natural variability of atomic movements.
Although \textit{Random Choice} improves diversity, it leads to worse FID because randomly selected movements may require stronger temporal scaling, making the completed motion less natural.
It also obtains a lower $R$, since excessive randomness may cause different atomic movements to be selected even under similar musical structures.
In comparison, the proposed \textit{Duration Nearest} strategy achieves the best overall balance between motion diversity, structural consistency, and rhythmic coherence.
Its higher MultiModality compared with \textit{Single Candidate} further confirms that duration-aware retrieval helps the completion stage generate diverse dances while preserving temporal compatibility.

\begin{figure}[t]
\centering
\includegraphics[width=1\textwidth]{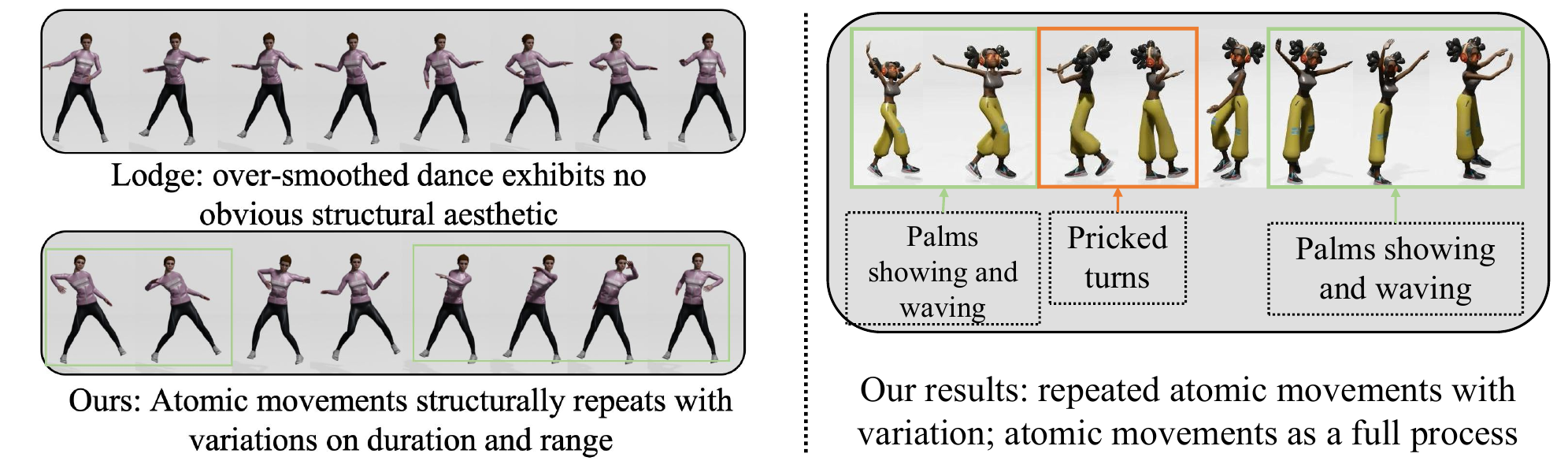}
\caption{\label{fig:vis_res} Qualitative results and comparisons with Lodge\cite{lodge}.
}
\end{figure}

\subsection{Qualitative Results}

On the left of \cref{fig:vis_res}, we qualitative compare our method with Lodge\cite{lodge}. We find that ours exhibits typical atomic movements repeatedly with variations on duration and range. However, Lodge suffers from an over-smooth dance with no apparent dance primitive. On the right, we show more results where the atomic movements show as a full process, exhibit repetition and variation, and are interpretable. 
We show more visualizations in the supplementary materials.
\section{Conclusions}

In this work, we present a structure-aware framework for music-to-dance generation that explicitly models dance as a sequence of semantically interpretable atomic movements. By decoupling the generation into atomic movement planning and motion completion, our method effectively mimics the 2-stage choreography workflow of human creation — planning and performance — achieving both structural coherence and expressive flexibility. Comprehensive experiments demonstrate it produces more musically aligned and structurally consistent dances and enables fine-grained interpretable control over composition.

\section*{Acknowledgment}
This work was supported by the grants from the Beijing Nova Program, Beijing Natural Science Foundation 4252040, National Natural Science Foundation of China 62372014 and CAAI-Tencent Rhino-Bird Open Research Fund.



%
%
\bibliographystyle{splncs04}
\bibliography{main}

@String(CVPR  = {IEEE Conf. Comput. Vis. Pattern Recog.})

@String(ICCV  = {Int. Conf. Comput. Vis.})

@String(ECCV  = {Eur. Conf. Comput. Vis.})

@String(CVPR  = {CVPR})

@String(ICCV  = {ICCV})

@String(ECCV  = {ECCV})

@inproceedings{bailando,
    title={Bailando: 3D dance generation via Actor-Critic GPT with Choreographic Memory},
    author={Siyao, Li and Yu, Weijiang and Gu, Tianpei and Lin, Chunze and Wang, Quan and Qian, Chen and Loy, Chen Change and Liu, Ziwei },
    booktitle={CVPR},
    year={2022}
}

@misc{tm2d,
      title={TM2D: Bimodality Driven 3D Dance Generation via Music-Text Integration}, 
      author={Kehong Gong and Dongze Lian and Heng Chang and Chuan Guo and Zihang Jiang and Xinxin Zuo and Michael Bi Mi and Xinchao Wang},
      year={2023},
      eprint={2304.02419},
      archivePrefix={arXiv},
      primaryClass={cs.CV},
      url = { }, 
}

@misc{aist,
      title={AI Choreographer: Music Conditioned 3D Dance Generation with AIST++}, 
      author={Ruilong Li and Shan Yang and David A. Ross and Angjoo Kanazawa},
      year={2021},
      eprint={2101.08779},
      archivePrefix={arXiv},
      primaryClass={cs.CV},
      url = { }, 
}

@inproceedings{
MDM,
title={Human Motion Diffusion Model},
author={Guy Tevet and Sigal Raab and Brian Gordon and Yoni Shafir and Daniel Cohen-or and Amit Haim Bermano},
booktitle={The Eleventh International Conference on Learning Representations },
year={2023},
url = { }
}

@ARTICLE{MD,
  author={Zhang, Mingyuan and Cai, Zhongang and Pan, Liang and Hong, Fangzhou and Guo, Xinying and Yang, Lei and Liu, Ziwei},
  journal={IEEE Transactions on Pattern Analysis and Machine Intelligence}, 
  title={MotionDiffuse: Text-Driven Human Motion Generation With Diffusion Model}, 
  year={2024},
  volume={46},
  number={6},
  pages={4115-4128},
  doi={ }}

@InProceedings{humanml3d,
    author    = {Guo, Chuan and Zou, Shihao and Zuo, Xinxin and Wang, Sen and Ji, Wei and Li, Xingyu and Cheng, Li},
    title     = {Generating Diverse and Natural 3D Human Motions From Text},
    booktitle = {Proceedings of the IEEE/CVF Conference on Computer Vision and Pattern Recognition (CVPR)},
    month     = {June},
    year      = {2022},
    pages     = {5152-5161}
}

@INPROCEEDINGS{edge,
  author={Tseng, Jonathan and Castellon, Rodrigo and Liu, C. Karen},
  booktitle={2023 IEEE/CVF Conference on Computer Vision and Pattern Recognition (CVPR)}, 
  title={EDGE: Editable Dance Generation From Music}, 
  year={2023},
  volume={},
  number={},
  pages={448-458},
  doi={ }}

@inproceedings{petrovich23tmr,
    title     = {{TMR}: Text-to-Motion Retrieval Using Contrastive {3D} Human Motion Synthesis},
    author    = {Petrovich, Mathis and Black, Michael J. and Varol, G{\"u}l},
    booktitle = {International Conference on Computer Vision ({ICCV})},
    year      = {2023}
}

@inproceedings{nam2021zero,
  title={Zero-shot Natural Language Video Localization},
  author={Nam, Jinwoo and Ahn, Daechul and Kang, Dongyeop and Ha, Seong Jong and Choi, Jonghyun},
  booktitle={Proceedings of the IEEE/CVF International Conference on Computer Vision},
  pages={1470-1479},
  year={2021}
}

@inproceedings{delmas2022posescript,
  title={{PoseScript: 3D Human Poses from Natural Language}},
  author={{Delmas, Ginger and Weinzaepfel, Philippe and Lucas, Thomas and Moreno-Noguer, Francesc and Rogez, Gr\'egory}},
  booktitle={{ECCV}},
  year={2022}
}

@inproceedings{t2mgpt,
  title={T2M-GPT: Generating Human Motion from Textual Descriptions with Discrete Representations},
  author={Zhang, Jianrong and Zhang, Yangsong and Cun, Xiaodong and Huang, Shaoli and Zhang, Yong and Zhao, Hongwei and Lu, Hongtao and Shen, Xi},
  booktitle={Proceedings of the IEEE/CVF Conference on Computer Vision and Pattern Recognition (CVPR)},
  year={2023},
}

@inproceedings{TEMOS,
  title     = {{TEMOS}: Generating diverse human motions from textual descriptions},
  author    = {Petrovich, Mathis and Black, Michael J. and Varol, G{\"u}l},
  booktitle = {European Conference on Computer Vision ({ECCV})},
  year      = {2022}
}

@inproceedings{ddpm,
 author = {Ho, Jonathan and Jain, Ajay and Abbeel, Pieter},
 booktitle = {Advances in Neural Information Processing Systems},
 editor = {H. Larochelle and M. Ranzato and R. Hadsell and M.F. Balcan and H. Lin},
 pages = {6840--6851},
 publisher = {Curran Associates, Inc.},
 title = {Denoising Diffusion Probabilistic Models},
 url = { },
 volume = {33},
 year = {2020}
}

@misc{gemini,
      title={Gemini: A Family of Highly Capable Multimodal Models}, 
      author={Gemini Team and Rohan Anil and Sebastian Borgeaud and Jean-Baptiste Alayrac and others},
      year={2024},
      eprint={2312.11805},
      archivePrefix={arXiv},
      primaryClass={cs.CL},
      url = { }, 
}

@article{B,
    author    = {Powell, H. E.},
    title     = {Modern Dance Choreography: Beyond the Movement an Analysis between Lyrics and Movement: Can identities be developed through modern dance choreography?},
    journal   = {Annual Review of Education, Communication \& Language Sciences},
    year      = {2019},
    volume    = {16},
    number    = {2},
}

@inproceedings{lodge,
  title={Lodge: A coarse to fine diffusion network for long dance generation guided by the characteristic dance primitives},
  author={Li, Ronghui and Zhang, YuXiang and Zhang, Yachao and Zhang, Hongwen and Guo, Jie and Zhang, Yan and Liu, Yebin and Li, Xiu},
  booktitle={Proceedings of the IEEE/CVF Conference on Computer Vision and Pattern Recognition},
  pages={1524--1534},
  year={2024}
}

@misc{beatit,
      title={Beat-It: Beat-Synchronized Multi-Condition 3D Dance Generation}, 
      author={Zikai Huang and Xuemiao Xu and Cheng Xu and Huaidong Zhang and Chenxi Zheng and Jing Qin and Shengfeng He},
      year={2024},
      eprint={2407.07554},
      archivePrefix={arXiv},
      primaryClass={cs.GR},
      url = { }, 
}

@INPROCEEDINGS{danceba,
      author={Fan, Congyi and Guan, Jian and Zhao, Xuanjia and Xu, Dongli and Lin, Youtian and Ye, Tong and Feng, Pengming and Pan, Haiwei},
  booktitle={2025 IEEE/CVF International Conference on Computer Vision (ICCV)}, 
  title={Align Your Rhythm: Generating Highly Aligned Dance Poses with Gating-Enhanced Rhythm-Aware Feature Representation}, 
  year={2025},
  volume={},
  number={},
  pages={13193-13202},
  doi={ }
    }

@inproceedings{
dancerevo,
title={ Dance Revolution: Long-Term Dance Generation with Music via Curriculum Learning},
author={Ruozi Huang and Huang Hu and Wei Wu and Kei Sawada and Mi Zhang and Daxin Jiang},
booktitle={International Conference on Learning Representations},
year={2021}
}

@article{dancenet,
author = {Zhuang, Wenlin and Wang, Congyi and Chai, Jinxiang and Wang, Yangang and Shao, Ming and Xia, Siyu},
title = {Music2Dance: DanceNet for Music-Driven Dance Generation},
year = {2022},
issue_date = {May 2022},
publisher = {Association for Computing Machinery},
address = {New York, NY, USA},
volume = {18},
number = {2},
issn = {1551-6857},
url = { },
doi = { },
journal = {ACM Trans. Multimedia Comput. Commun. Appl.},
month = feb,
articleno = {65},
numpages = {21}
}

@inproceedings{D3PM,
author = {Austin, Jacob and Johnson, Daniel D. and Ho, Jonathan and Tarlow, Daniel and van den Berg, Rianne},
title = {Structured denoising diffusion models in discrete state-spaces},
year = {2021},
isbn = {9781713845393},
publisher = {Curran Associates Inc.},
address = {Red Hook, NY, USA},
booktitle = {Proceedings of the 35th International Conference on Neural Information Processing Systems},
articleno = {1376},
numpages = {13},
series = {NIPS '21}
}

@ARTICLE{atomic1,
  title    = "A primitive-based representation of dance: modulations by
              experience and perceptual validity",
  author   = "Leh, Amalaswintha and Endres, Dominik and Hegele, Mathias",
  journal  = "J Neurophysiol",
  volume   =  130,
  number   =  5,
  pages    = "1214--1225",
  month    =  oct,
  year     =  2023,
  address  = "United States",
  language = "en"
}

@inproceedings{atomic2,
author = {Okada, N. and Iwamoto, Naoya and Fukusato, Tsukasa and Morishima, Shigeo},
year = {2015},
month = {03},
pages = {332-339},
title = {Dance motion segmentation method based on choreographic primitives},
journal = {GRAPP 2015 - 10th International Conference on Computer Graphics Theory and Applications; VISIGRAPP, Proceedings},
doi = { }
}

@article{atmoic0,
title = {Fractal patterns in music},
journal = {Chaos, Solitons \& Fractals},
volume = {170},
pages = {113315},
year = {2023},
issn = {0960-0779},
doi = { },
url = { },
author = {John McDonough and Andrzej Herczyński}
}

@article{atmoic3,
author = {Darrell Conklin},
title = {Pattern in music},
journal = {Journal of Mathematics and Music},
volume = {15},
number = {2},
pages = {95--98},
year = {2021},
publisher = {Taylor \& Francis},
doi = { },


URL = { },
eprint = { 


}

}

@article{atomicstruct,
author = {Reese, Spencer},
year = {2019},
month = {04},
pages = {},
title = {Music in Motion: Interpreting Musical Structure through Choreography in Gershwin’s "An American in Paris"}
}

@inproceedings{edmg,
author = {Zhang, Jinming and Sun, Yunlian and Zhang, Hongwen and Tang, Jinhui},
title = {EDMG: Towards Efficient Long Dance Motion Generation with Fundamental Movements from Dance Genres},
year = {2025},
isbn = {9798400720352},
publisher = {Association for Computing Machinery},
address = {New York, NY, USA},
url = { },
doi = { },
booktitle = {Proceedings of the 33rd ACM International Conference on Multimedia},
pages = {10447–10456},
numpages = {10},
location = {Dublin, Ireland},
series = {MM '25}
}

@misc{gen1,
      title={Weakly Supervised Dynamic Scene Graph Generation with Temporal-enhanced In-domain Knowledge Transferring}, 
      author={Zhu Xu and Ting Lei and Zhimin Li and Guan Wang and Qingchao Chen and Yuxin Peng and Yang Liu},
      year={2025},
      booktitle={ICCV},
      organization={IEEE}
}

@inproceedings{gen2,
author = {Xu, Zhu and Wang, Zhaowen and Peng, Yuxin and Liu, Yang},
title = {Interact-Custom: Customized Human Object Interaction Image Generation},
year = {2025},
isbn = {9798400720352},
publisher = {Association for Computing Machinery},
address = {New York, NY, USA},
url = { },
doi = { },
booktitle = {Proceedings of the 33rd ACM International Conference on Multimedia},
pages = {9500–9508},
numpages = {9},
location = {Dublin, Ireland},
series = {MM '25}
}

@inproceedings{interact,
author = {Cai, Xinhao and Zheng, Minghang and Jin, Xin and Liu, Yang},
title = {InteractMove: Text-Controlled Human-Object Interaction Generation in 3D Scenes with Movable Objects},
year = {2025},
isbn = {9798400720352},
publisher = {Association for Computing Machinery},
address = {New York, NY, USA},
url = { },
doi = { },
booktitle = {Proceedings of the 33rd ACM International Conference on Multimedia},
pages = {9491–9499},
numpages = {9},
location = {Dublin, Ireland},
series = {MM '25}
}

@Article{poseEst,
title = {FMR-GNet: Forward Mix-Hop Spatial-Temporal Residual Graph Network for 3D Pose Estimation},
journal = {Chinese Journal of Electronics},
volume = {33},
number = {6},
pages = {1346-1359},
year = {2024},
issn = {},
doi = { },	
url = { },
author = {YANG Honghong and LIU Hongxi and ZHANG Yumei and WU Xiaojun}
}

@ARTICLE{danceUnderstand,
  author={Yang, Honghong and Wang, Sai and Jiang, Lu and Zhang, Yumei and Wu, Xiaojun},
  journal={Chinese Journal of Electronics}, 
  title={DTI-GNet: Dynamic Topology Inferenced Graph Convolution Network for Dance Action Recognition}, 
  year={2025},
  volume={34},
  number={4},
  pages={1284-1299},
  doi = { }}

@ARTICLE{mm1,
  author={Yang, Jie and Ma, Miao and Li, Yutong and Pei, Zhao},
  journal={Chinese Journal of Electronics}, 
  title={VQALS: A Video Question Answering Method in Low-Light Scenes Based on Illumination Correction and Feature Enhancement}, 
  year={2025},
  volume={34},
  number={4},
  pages={1300-1308},
  doi={ }}

@ARTICLE{mm2,
  author={Xu, Huiyu and Jin, Shuaifan and Wang, Zhibo and Ba, Zhongjie and Wei, Tao},
  journal={Chinese Journal of Electronics}, 
  title={PSA-NeRF: Personalized Spatial Attention Neural Rendering for Audio-Driven Talking Portraits Generation}, 
  year={2025},
  volume={34},
  number={6},
  pages={1821-1831},
  doi={ }}
\end{document}